\documentclass{article}

\usepackage{ijcai18}

\usepackage{times}
\usepackage{soul}
\usepackage{url}
\usepackage[hidelinks]{hyperref}
\usepackage[utf8]{inputenc}
\usepackage[small]{caption}
\usepackage[belowskip=0pt,aboveskip=5pt]{caption}
\usepackage{subcaption}

\usepackage{tasks}

\usepackage{xcolor}
\usepackage{graphicx}

\usepackage{multirow}
\usepackage{float}

\usepackage{algorithm}
\usepackage[noend]{algpseudocode}

\usepackage{amsmath}
\usepackage{algorithmicx}

\usepackage[bottom]{footmisc}
\usepackage{titlesec}

\makeatletter
\def\BState{\State\hskip-\ALG@thistlm}
\makeatother

\title{Goal-Oriented Chatbot Dialog Management Bootstrapping with Transfer Learning}

\author{
Vladimir Ilievski$^1$, 
Claudiu Musat$^2$, 
Andreea Hossmann$^2$,
Michael Baeriswyl$^2$
\\
$^1$ School of Computer and Communication Sciences, EPFL, Switzerland \\
$^2$ Artificial Intelligence Group - Swisscom AG
\\
Vladimir.Ilievski@epfl.ch, Claudiu.Musat@swisscom.com, \\ Andreea.Hossmann@swisscom.com, Michael.Baeriswyl@swisscom.com
}

\begin{document}

\maketitle

\begin{abstract}
Goal-Oriented (GO) Dialogue Systems, colloquially known as goal oriented chatbots, help users achieve a predefined goal (e.g. book a movie ticket) within a closed domain. A first step is to understand the user's goal by using natural language understanding techniques. Once the goal is known, the bot must manage a dialogue to achieve that goal, which is conducted with respect to a learnt policy. 
The success of the dialogue system depends on the quality of the policy, which is in turn reliant on the availability of high-quality training data for the policy learning method, for instance Deep Reinforcement Learning. 

Due to the domain specificity, the amount of available data is typically too low to allow the training of good dialogue policies. In this paper we introduce a transfer learning method to mitigate the effects of the low in-domain data availability. Our transfer learning based approach improves the bot's success rate by $20\%$ in relative terms for distant domains and we more than double it for close domains, compared to the model without transfer learning. Moreover, the transfer learning chatbots learn the policy up to 5 to 10 times faster. Finally, as the transfer learning approach is complementary to additional processing such as warm-starting, we show that their joint application gives the best outcomes.

\end{abstract}

\section{Introduction}

Text-based Dialogue Systems, colloquially known as chatbots, are widely used today in a plethora of different applications, ranging from trivial chit-chatting to personal assistants. Depending on the nature of the conversation, the Dialogue Systems can be classified in $i)$ open-domain~\cite{serban2016building,vinyals2015neural} and $ii)$ closed-domain Dialogue Systems~\cite{wen2016network}.
Goal-Oriented (GO) Chatbots are designed to help users to achieve predetermined goals (e.g. book a movie ticket)~\cite{peng2017composite}. These bots are closed-domain and can be grouped together in larger systems such as Amazon Alexa\footnote{https://developer.amazon.com/alexa} to give the impression of a general coverage. Each individual component (which in Amazon Alexa can be viewed as skills of the overarching generalist bot) is closed-domain in nature. 

The availability of data within a closed domain poses a major obstacle in the development of useful GO dialogue systems. Not all systems have the same data requirements. There are two dominant paradigms in Goal-Oriented (GO) Dialogue Systems implementations. The first type are fully supervised, sequence-to-sequence~\cite{sutskever2014sequence} models, that encode a user request and its context and decode a bot answer directly.  The fully supervised Goal-Oriented chatbots require a considerable amount of annotated dialogues, because they mimic the knowledge of the expert~\cite{wen2016network}.

\begin{figure}[t!]
\centering
\hbox{\hspace{-0.4em}\includegraphics[width=0.49\textwidth]{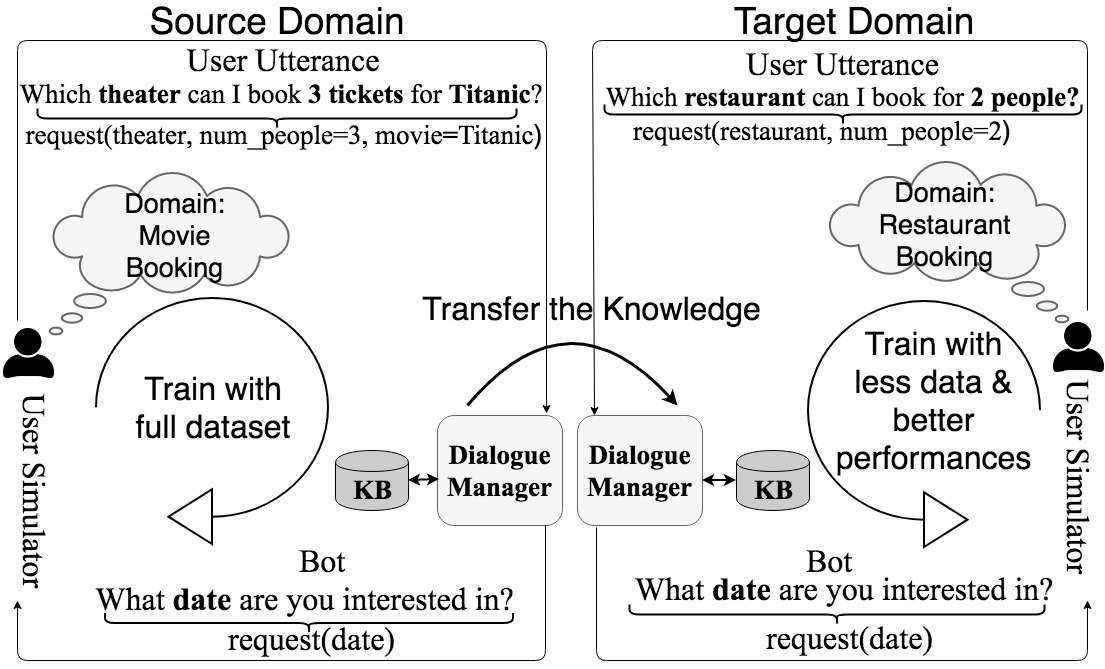}}
\caption{Model of the Goal-Oriented Dialogue System operating on a semantic level. We transfer the in-domain knowledge.}
\label{fig:model}
\end{figure}

A second type are algorithms based on reinforcement learning (RL), for instance based on Deep Q-Nets (DQN)~\cite{mnih2015human}. The DQN technique successfully applies a supervised learning methodology in the task of reinforcement learning. In their work, ~\cite{li2017end} use it to successfully build a Goal-Oriented Dialogue System in the domain of movie booking.
The lack of in-domain dialogue data is a key problem for training high quality DQN-based Goal-Oriented chatbots. We need in-domain labeled dialogues for two reasons: $i)$ to warm-start the chatbot, which is a standard widely used technique and $ii)$ to train the chatbot by simulating a considerable number of different conversations.

RL bots require less annotated dialogues than their sequence-to-sequence counterparts, due to their ability to simulate the conversation, thus exploring the unknown dialogue space more efficiently. The data requirements are however not trivial and obtaining the dialogue data is still the biggest obstacle their creators face.

In this paper, we show that we can build better Goal-Oriented Dialogue Systems using \textit{Transfer Learning}. We leverage the similarity between a source and a target domain, as many domains, such as restaurant and movie booking, share to a large extent some common information. 
In earlier work, bots were created independently for each domain (e.g. ~\cite{li2017end} created a bot for movie booking and ~\cite{wen2016network} one for restaurant reservations). These two domains include information for time and place. We believe this information need not be learnt twice and that a transfer is possible.
We show this possible connection graphically in Figure \ref{fig:model}.

We distinguish two cases: $i)$ when the two domains have an overlap and $ii)$ when one domain is an extension of another. To the best of our knowledge we are the first ones to successfully use and combine the Transfer Learning and the Goal-Oriented Chatbots based on the Deep Reinforcement Learning techniques. The contributions of this work are the following:

\begin{itemize}
\item \textbf{Training GO chatbots with less data}: In data constrained environments, models trained with transfer learning achieve better training and testing performances than ones trained independently.

\item \textbf{Better GO chatbot performance}: Using transfer learning has a significant positive effect on performance even when all the data from the target domain is available.

\item \textbf{Intuitions on further improvements}: We show the gains obtained with transfer learning are complementary to the ones due to warm-starting and the two can be successfully combined. 

\item \textbf{New published datasets}: We publish new datasets for training Goal-Oriented Dialogue Systems, for restaurant booking and tourist info domains\footnote{\href{https://github.com/IlievskiV/Master_Thesis_GO_Chatbots}{https://github.com/IlievskiV/Master\_Thesis\_GO\_Chatbots}}. They are derived from the third Dialogue State Tracking Challenge~\cite{henderson2013dialog}.
\end{itemize}

The rest of the paper is organized as follows: Section \ref{sec:related_work} presents the related work for Goal-Oriented Dialogue Systems based on Deep Reinforcement Learning techniques and for bots in data-constrained environments. Our model is fully detailed in Section \ref{sec:model}. We further describe the use of Transfer Learning technique in Section \ref{sec:transfer_learning}. We conduct our experiments and show the results in Section \ref{sec:experiments}. Finally, we conclude our work in Section \ref{sec:conclusion}.

\section{Related Work}
\label{sec:related_work}

\subsection{Goal-oriented Dialogue Systems}

The Goal-Oriented Dialogue Systems have been under development in the past two decades, starting from the basic, handcrafted Dialogue Systems~\cite{zue2000juplter}. The recent efforts to build such systems are generally divided in two lines of research. 

The first way is to treat them in an end-to-end, fully supervised manner, in order to use the power of the deep neural networks based on the encoder-decoder principle to infer the latent representation of the dialogue state. The authors in ~\cite{vinyals2015neural} used standard Recurrent Neural Networks (RNNs) and trained a Goal-Oriented Chatbot in a straightforward sequence-to-sequence~\cite{sutskever2014sequence} fashion. On the other hand, ~\cite{serban2016building} utilized a hierarchical RNNs to do the same task. Additionally, in their work~\cite{bordes2016learning} used the memory networks~\cite{sukhbaatar2015end} to build a Goal-Oriented Chatbot for restaurant reservation.

Another branch of research had emerged, focusing on the Deep Reinforcement Learning because the supervised approach is data-intensive. These techniques require less annotated data because of their sequential nature to simulate the dialogue and explore different aspects of the dialogue space. In their work~\cite{li2017end,dhingra2016end,cuayahuitl2017simpleds} successfully applied the Deep Reinforcement Learning combined with a user simulator to build GO Dialogue Systems.

However, these models are quite complex since they include many submodules, such as Natural Language Understanding (NLU)~\cite{hakkani2016multi} and Natural Language Generation (NLG)~\cite{wen2015semantically} units, as well as a Dialogue State Tracker (DST), which introduce significant noise. For this reason, there is a line of research that combined both approaches. In their work ~\cite{su2016continuously} first trained the policy network in a supervised learning fashion and then fine-tuned the policy using the Reinforcement Learning.

\subsection{Data-constrained Dialogue Systems}

One desired property of the Goal-Oriented Dialogue Systems is the ability to switch to new domains and at the same time not to lose any knowledge learned from training on the previous ones. In this direction, the authors in ~\cite{gavsic2015distributed} proposed a Gaussian Process-based technique to learn generic dialogue polices. These policies with a little amount of data can be furthermore adjusted according to the use case of the dialogue system. On the other hand,~\cite{wang2015learning} learned domain-independent dialogue policies, such that they parametrized the ontology of the domains. In this way, they show that the policy optimized for a restaurant search domain can be successfully deployed to a lap-top sale domain. Last but not the least, ~\cite{lee2017toward} utilized a continual learning, to smoothly add new knowledge in the neural networks that specialized a dialogue policy in an end-to-end fully supervised manner.  

Nevertheless, none of the previously mentioned papers tackles the problem of transferring the domain knowledge in case when the dialogue policy is optimized using a Deep Reinforcement Learning. In our work, we propose such method, based on the standard Transfer Learning technique~\cite{pan2010survey}. Therefore, using this method we surpass the limitations to transfer the in-domain knowledge in Goal-Oriented Dialogue Systems based on the Deep RL.

\section{Transfer Learning for Goal-Oriented Chatbots}
\label{sec:model}

Our primary goal is to use transfer learning to increase a chatbot's success rate. The success rate is the fraction of successfully conducted dialogues. A successful dialogue is one where the user gets a satisfactory answer before the maximum number of dialogue turns is exhausted. 

\subsection{Model}

Our work is based on the model from~\cite{li2017end}, who proposed an end-to-end reinforcement learning approach. ~\cite{li2017end} use an agenda-based user simulator to build a Goal-Oriented Chatbot in a movie booking domain.

Goal-oriented bots contain an initial natural understanding (NLU) component, that is tasked with determining the user's \textit{intent} (e.g. \textit{book a movie ticket}) and its parameters, also known as \textit{slots} (e.g. date: \textit{today}, count: \textit{three people}, time: \textit{7 pm}).
The usual practice in the RL-based Goal-Oriented Chatbots is to define the user-bot interactions as \textit{semantic frames}. At some point $\mathbf{t}$ in the time, given the user utterance $\mathbf{u_{t}}$, the system needs to perform an action $\mathbf{a_{t}}$. A bot action is, for instance, to request a value for an empty slot or to give the final result.

The entire dialogue can be reduced to a set of \textit{slot-value} pairs, called \textit{semantic frames}. 
Consequently, the conversation can be executed on two distinct levels:
\begin{itemize}
	\item \textbf{Semantic level:} the user sends and receives only a semantic frames as messages.
    \item \textbf{Natural language level:} the user sends and receives natural language sentences, which are reduced to, or derived from a semantic frame by using Natural Language Understanding (NLU) and Natural Language Generation (NLG) units respectively~\cite{wen2015semantically,hakkani2016multi}.
\end{itemize}

The composition of the Dialogue System we are using is shown in Figure \ref{fig:model}. It consists of two independent units: the \textit{User Simulator} on the left side and the \textit{Dialogue Manager} (DM) on the right side. We operate on the semantic level, removing the noise introduced by the NLU and NLG units. We want to focus exclusively on the impact of transfer learning techniques on dialog management.

\begin{figure*}[h!]
\centering
\begin{subfigure}[b]{.5\textwidth}
\centering
\includegraphics[width=\textwidth]{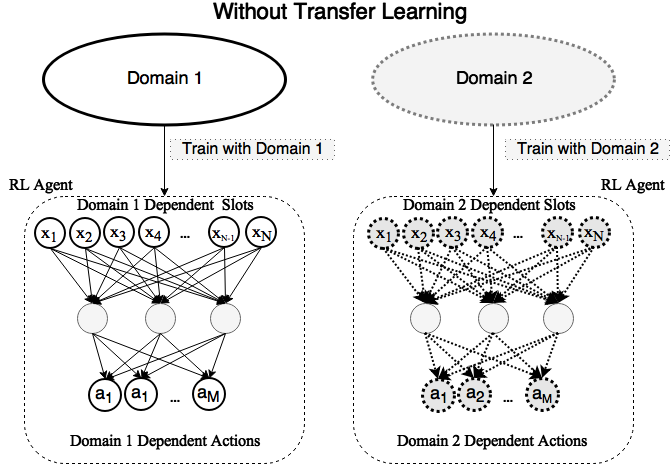}
\caption{The training process without transfer learning}
\label{subfig:no_transfer_learning}
\end{subfigure}%
\begin{subfigure}[b]{.5\textwidth}
\centering
\includegraphics[width=\textwidth]{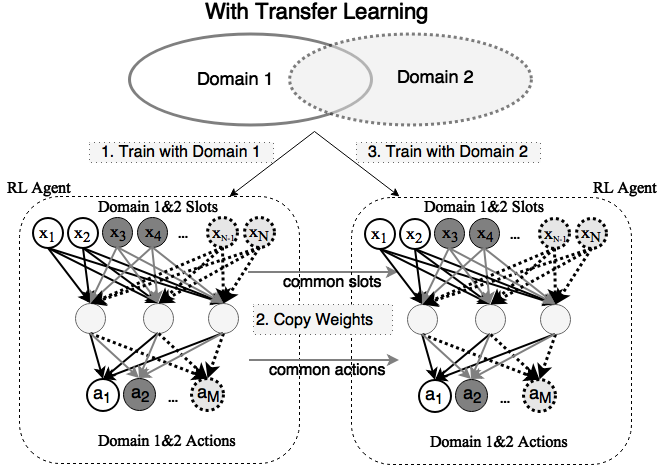}
\caption{The training process with transfer learning}
\label{subfig:transfer_learning}
\end{subfigure}
\caption{Comparison of the Goal-Oriented Dialogue System training process, without transfer learning (left side) and with transfer learning (right side).}
\label{fig:no_transfer_learning_vs_transfer_learning}
\end{figure*}

\subsection{User Simulator}

The User Simulator creates a user - bot conversation, given the semantic frames. Because the model is based on Reinforcement Learning, a dialogue simulation is necessary to successfully train the model. The user simulator we use in this work is based on the work by~\cite{li2016user}.

From the dataset of available user goals the Simulator randomly picks one, which is unknown for the Dialogue Manager. The user goal consists of two different sets of slots: \textit{inform slots} and \textit{request slots}. 
\begin{itemize} 
\item \textit{Inform slots} are the slots for which the user knows the value, i.e. they represent the user constraints (e.g. \{movie\_name:  ``avengers'', number\_of\_people:  ``3'', date:  ``tomorrow''\}). 
\item \textit{Request slots} are ones for which the user is looking for an answer (e.g. \{ city, theater, start\_time \} \}). 
\end{itemize}
Having the user goal as an anchor, the user simulator generates the \textit{user utterances} $\mathbf{u_{t}}$. The initial user utterance, similar to the user goal, consists of the initial inform and request sets of slots.
Additionally, it includes a user intent, like \textit{open dialogue} or \textit{request additional info}.

The user utterances generated over the course of the conversation follow an agenda-based model~\cite{schatzmann2009hidden}. According to this model, the user is having an internal state $\mathbf{s_{u}}$, which consists a goal $G$ and an agenda $A$. The goal furthermore is split in user constraints $C$ and user requests $R$. In every consecutive time step $\mathbf{t}$, the user simulator creates the user utterance $\mathbf{u_{t}}$, using its current state $\mathbf{s_{u}}$ and the last system action $\mathbf{a_{t}}$. In the end, using the newly generated user utterance $\mathbf{u_{t}}$, it updates the internal state $\mathbf{s^{\prime}_{u}}$.

\subsection{Dialogue Manager}
The Dialogue Manager (DM), as its name suggests, manages the dialogue flow in order to conduct a proper dialogue with the user. 
The DM is composed by two trainable subcomponents: the \textit{Dialogue State Tracker} (DST) and the \textit{Policy Learning Module}, i.e. the agent. Additionally, the Dialogue Manager exploits an external \textit{Knowledge Base} (KB), to find and suggest values for the user requests. Therefore, it plays a central role in the entire Dialogue System.

\subsubsection*{Dialogue State Tracker}
The responsibility of the Dialogue State Tracker (DST) is to build a reliable and robust representation of the current state of the dialogue. All system actions are based on the current dialogue state. It keeps track of the history of the user utterances, system actions and the querying results from the Knowledge Base. It extracts features and creates a vector embedding of the current dialogue state, which is exposed and used by the Policy Learning module later on. In order to produce the embeddings, the Dialogue State Tracker must know the type of all slots and intents that might occur during the dialogue. Since we operate on a semantic level (i.e. not introducing any additional noise), we employ a rule-based state tracker as in~\cite{li2017end}.

\subsubsection*{Policy Learning}
The Policy Learning module selects the next system actions to drive the user towards the goal in the smallest number of steps. It does that by using the deep reinforcement neural networks, called Deep Q-Networks (DQN)~\cite{mnih2015human}. DQNs successfully approximate the state-action function $Q\left( s, a | \theta \right)$ - a standard metric in the Reinforcement Learning, with latent parameters $\theta$. $Q\left( s, a | \theta \right)$ is the utility of taking an action $a$, when the agent is perceiving the state $s$, by following a policy $\pi = P(a | s)$. The utility measure is defined as the problem of maximizing the cumulative future reward that the agent will receive. 

DQNs contain a biologically inspired mechanism, called \textit{experience replay}. They store the agent's experiences $e_{t} = \left( s_{t}, a_{t}, r_{t}, s_{t+1} \right)$ in an \textit{experience replay buffer} $D_{t} = \left\{e_{1}, \cdots , e_{t}\right\}$, thus creating mini-batches of experiences, uniformly drawn from $D_{t}$ used to train the neural net.

In our case, the agent is getting the new state $s_{t}$ from the Dialogue State Tracker (DST) and then it takes a new action $a_{t}$, based on the $\epsilon$-greedy policy. It means, with a probability $\epsilon \in \left[0,1\right]$ it will take a random action, while with a probability $1 - \epsilon$ it will take the action resulting with a maximal Q-value. We thus trade between the exploration and exploitation of the dialogue space. 
For each slot that might appear in the dialogue, the agent can take two actions: either to ask the user for a constraining value or to suggest to the user a value for that slot. Additionally, there is a fixed size of slot-independent actions, to open and close the conversation. 

The agent receives positive and negative rewards accordingly, in order to force the agent to successfully conduct the dialogue. It is \textit{successful} if the number of totally required dialogue turns to reach the goal is less than a predefined maximal threshold $n_{max\_turns}$. For every additional dialogue turn, the agent receives a predefined negative reward $r_{ongoing}$. In the end, if the dialogue fails, it will receive a negative reward $r_{negative}$ equal to the negative of the predefined maximal allowed dialogue turns. If the dialogue is successful, it will receive a positive reward $r_{positive}$, two times the maximal allowed dialogue turns.

An important addition is the \textit{warm-starting} technique that fills the experience replay buffer with experiences coming from a successfully finished dialogues i.e. with positive experiences. 
This dramatically boosts the agent's performances before the actual training starts, as will be shown in Section \ref{sec:warmstart}. 
The training process continues with running a fixed number of independent training epochs $n_{epochs}$. In each epoch we simulate a predefined number of dialogues $n_{dialogues}$, thus filling the experience memory buffer. The result consists of mini-batches to train the underlying Deep Q-Net.

During the training process, when the agent reaches for the first time a success rate greater or equal to the success rate of a rule-based agent $s_{rule\_based}$, we flush the experience replay buffer, as described in detail in~\cite{li2017end}. 
 
\section{Transfer Learning}
\label{sec:transfer_learning}

The main goal of this work is to study the impact of a widely used technique - \textit{Transfer Learning} on goal oriented bots.
As the name suggests, transfer learning transfers knowledge from one neural network to another. The former is known as the source, while the latter is the target~\cite{pan2010survey}. The goal of the transfer is to achieve better performance on the target domain with limited amount of training data, while benefiting from additional information from the source domain. In the case of dialogue systems, the input space for both source and target nets are their respective dialogue spaces. 

The training process without transfer learning, shown in Figure \ref{subfig:no_transfer_learning}, processes the two dialogue domains independently, starting from randomly initialized weights. The results are dialogue states from separate distributions. Additionally, the sets of actions the agents might take in each domain are also independent.

\begin{algorithm}[b!]
\caption{Transfer Learning Pseudocode}
\label{alg:transfer_learning}

\begin{algorithmic}[1]
\Procedure{InitializeWeights}{sourceWeights, commonSlotIndices, commonActionIndices}

\State $ targetWeigths \gets \textit{RandInit()} $

\For{$i$ in $\textit{commonSlotIndices}$}
   \State $ \textit{targetWeigths} \left[ i \right] \gets sourceWeights \left[ i \right] $
\EndFor

\For{$i$ in $\textit{commonActionIndices}$}
   \State $ \textit{targetWeigths} \left[ i \right] \gets sourceWeights\left[ i\right ]$
\EndFor

\State \textbf{return} \textit{targetWeigths}
\EndProcedure
\end{algorithmic}

\end{algorithm}

On the other hand, as depicted in Figure \ref{subfig:transfer_learning} if we want to benefit from transfer learning, we must model the dialogue state in both domains, as if they were coming from the same distribution. The sets of actions have to be shared too. The bots specialized in the source domain must be aware of the actions in the second domain, even if these actions are never used, and vice versa. This requirement stems from the impossibility of reusing the neural weights if the input and output spaces differ.
Consequently, when we train the model on the source domain, the state of the dialogue depends not only on the slots that are specific to the source, but also on those that only appear in the target one. This insight can be generalized to a plurality of source and target domains. The same holds for the set of actions. 

When training the target domain model, we no longer randomly initializing all weights. The weights related to the source domain - both for slots and actions - are copied from the source model. The pseudocode for this weight initialization is portrayed in the Algorithm \ref{alg:transfer_learning}.

\section{Experiments}
\label{sec:experiments}

All experiments are executed using a setup template. Firstly, we train a model on the \textit{source domain} and reuse the common knowledge to boost the training and testing performance of the model trained on a different, but similar \textit{target domain}. Secondly, we train a model exclusively on the target domain, without any prior knowledge. This serves as a baseline. Finally, we compare the results of these two models. We thus have two different cases:

\begin{figure}[h!]
\centering
\includegraphics[width=.4\textwidth]{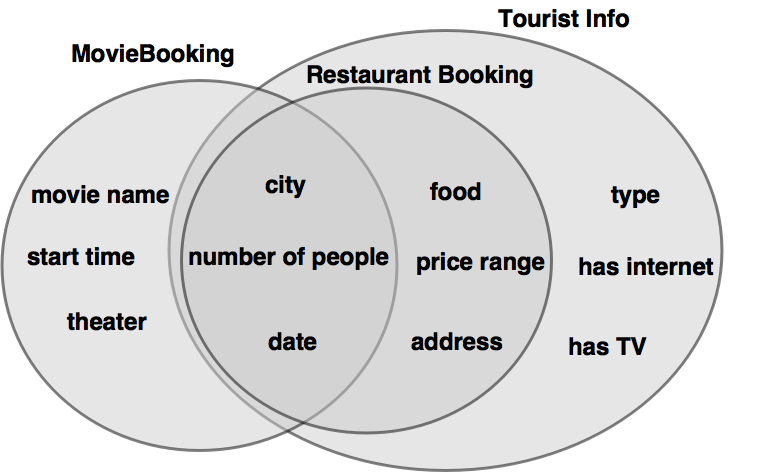}
\caption{Slot types in the three different domains}
\label{fig:slot_type}
\end{figure}

\begin{enumerate}
	\item \textit{Domain Overlap} - the source \(Movie Booking\) and target \(Restaurant Booking\) domains are different, but share a fraction of the slots.
    \item \textit{Domain Extension} - the source domain, now \(Restaurant Booking\), is extended to Tourist Information, that  contains all the slots from the source domain along with some additional ones.
\end{enumerate}

The reason for the choice of source domain in the domain overlap case is designed to enable a direct comparison to the results of ~\cite{li2017end}, who built a GO bot for movie booking. For the domain extension case, the only combination available was \(Restaurant - Tourism\).
The type of slots in each domain are given in Figure \ref{fig:slot_type}. For each domain, we have a training set of 120 user goals, and a testing set of 32 user goals.

Following the above mentioned setup template, we conduct two sets of experiments for each of the two cases. The first set shows the overall performance of the models leveraging the transfer learning approach. The second set shows the effects of the warm-starting jointly used with the transfer learning technique.

In all experiments, when we use a warm-starting, the criterion is to fill agent's buffer, such that 30 percent of the buffer is filled with positive experiences (coming from a successful dialogue). After that, we train for $n_{epochs} = 50$ epochs, each simulating $n_{dialogues} = 100$ dialogues. We flush the agent's buffer when the agent reaches, for a first time, a success rate of $s_{rule\_based} = 0.3$. We set the maximal number of allowed dialogue turns $n_{max\_turns}$ to 20, thus the negative reward $r_{negative}$ for a failed dialogue is $-20$, while the positive reward $r_{positive}$ for a successful dialogue is $40$. In the consecutive dialogue turns over the course of the conversation, the agent receives a negative reward of $r_{ongoing} = -1$. In all cases we set $\epsilon = 0.05$ to leave a space for exploration. By using this hyperparameters, we prevent the system to overfit and to generalize very well over the dialogue space. Finally, we report the success rate as a performance measure.

\subsection{Training GO Bots with Less Data}

Due to labeling costs, the availability of in-domain data is the bottleneck for training successful and high performing Goal-Oriented chatbots. We thus study the effect of transfer learning on training bots in data-constrained environments.

From the available 120 user goals for each domain's training set, we randomly select subsets of 5, 10, 20, 30, 50 and all 120. We then warm-start and train both the independent and transfer learning models on these sets. We test the performance on both the training set (\textit{training performance}) and the full set of 32 test user goals (\textit{testing performance}). We repeat the same experiment 100 times, in order to reduce the uncertainty introduced by the random selection. Finally, we report the success rate over the user goal portions with a 95\% confidence interval.

\begin{figure}[b!]
\centering
\begin{subfigure}[t]{.5\textwidth}
\centering
\includegraphics[width=.49\textwidth]{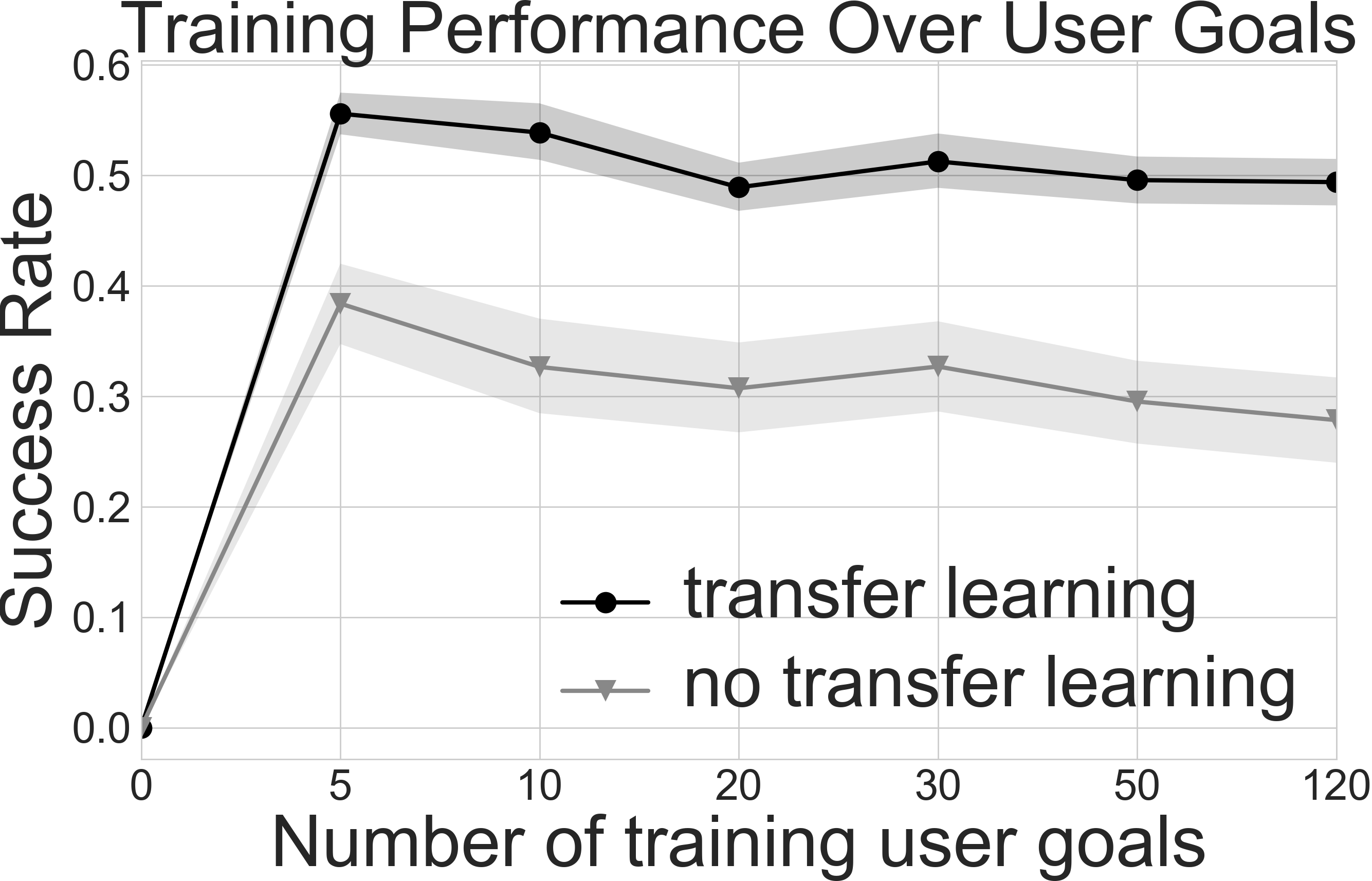}
\includegraphics[width=.49\textwidth]{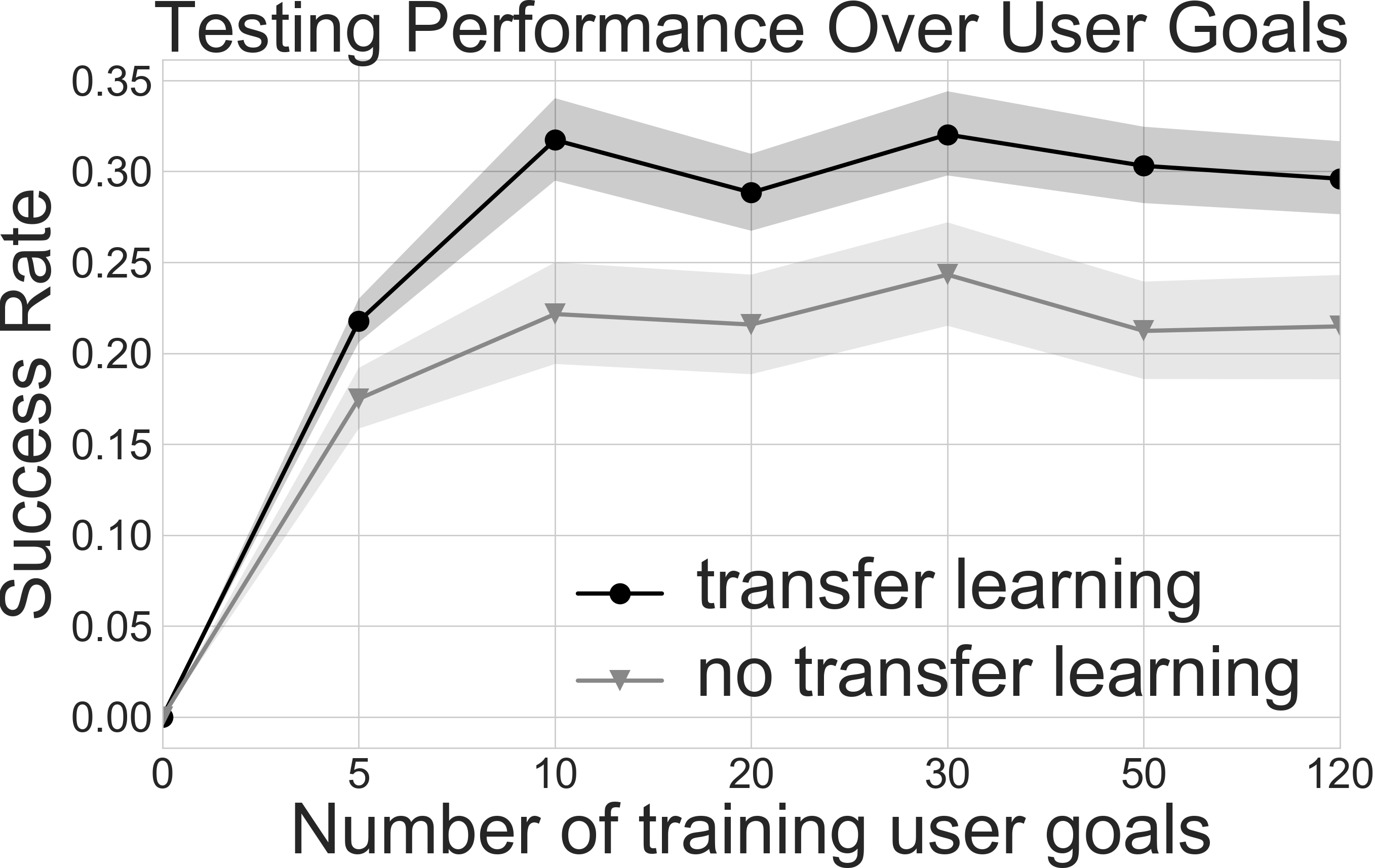}
\caption{Restaurant Booking with pre-training on Movie Booking domain}
\label{subfig:warm_up_rest_booking}
\end{subfigure}\\%
\begin{subfigure}[b]{.49\textwidth}
\centering
\includegraphics[width=.49\textwidth]{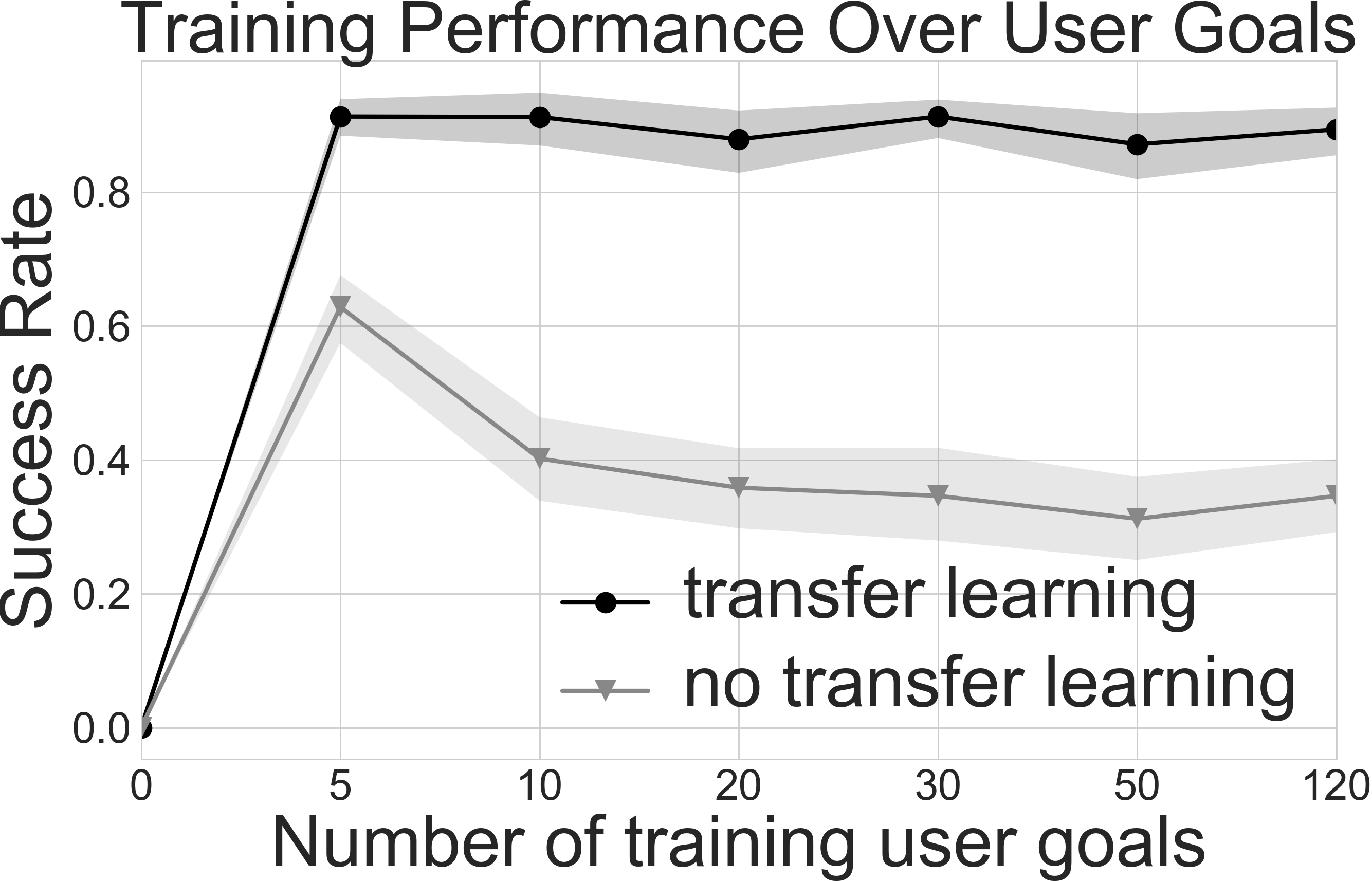}
\includegraphics[width=.5\textwidth]{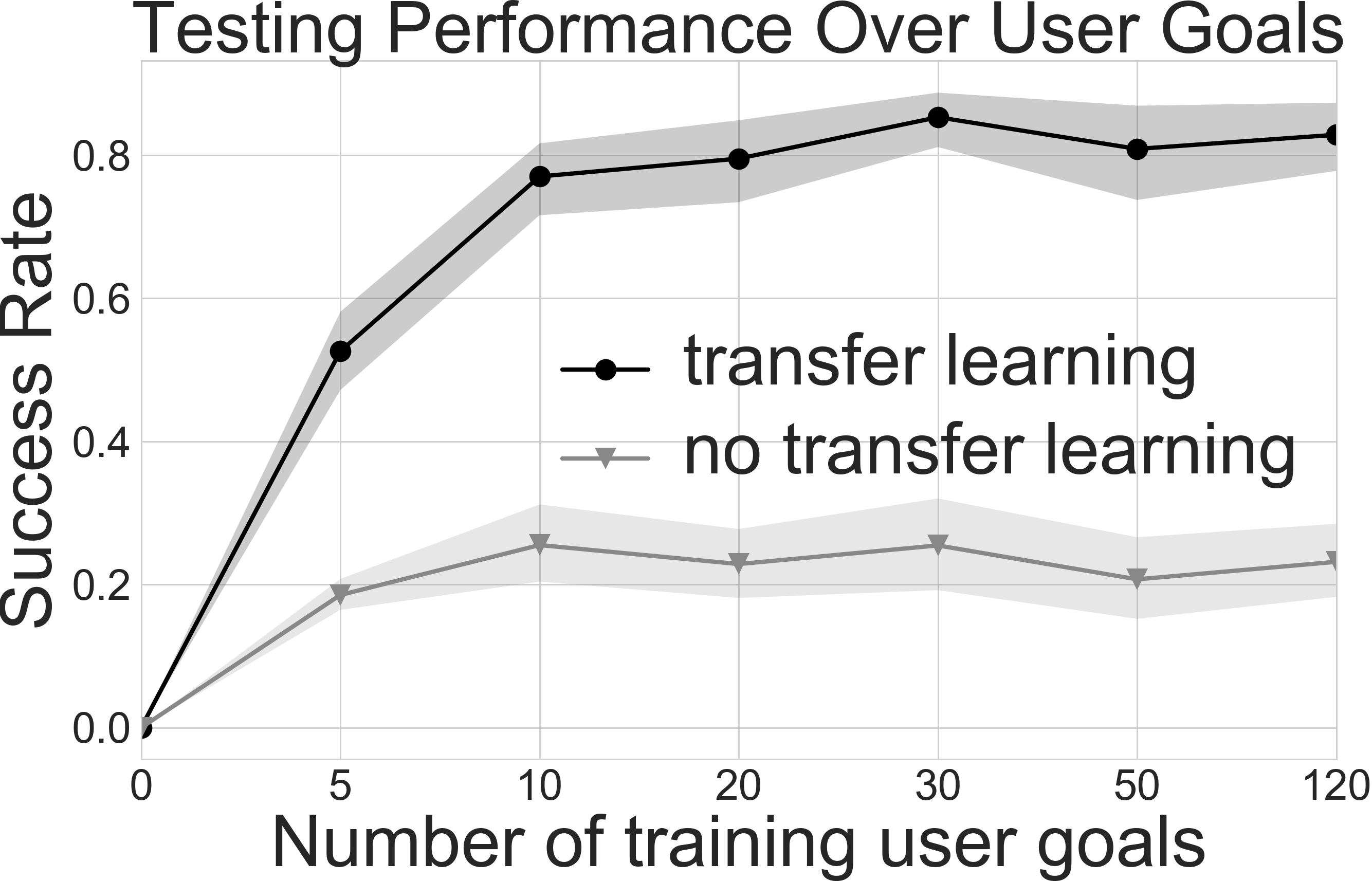}
\caption{Tourist Info with pre-training on Restaurant Booking domain}
\label{subfig:warm_up_tourist_info}
\end{subfigure}
\caption{Average training and testing success rates with 95\% confidence, for 100 runs over a randomly selected user goal portions of size 5, 10, 20, 30, 50 and 120, for both models: with and without transfer learning.}
\label{fig:warming_up_user_goal_portions}
\end{figure}

The training and testing results, in the first case of domain overlapping, are shown in Figure \ref{subfig:warm_up_rest_booking}. The success rate of the model obtained with transfer learning is 65\% higher than that of the model trained without any external prior knowledge. In absolute terms the success rate climbs on average from 30\% to 50\%. For the test dataset, transfer learning improves the success rate from 25\% to 30\%, for a still noteworthy 20\% relative improvement.

In the case of domain extension, the difference between the success rates of the two models is even larger (Figure \ref{subfig:warm_up_tourist_info}). This was expected, as the extended target domain contains all slots from the source domain, therefore not losing any source domain information. The overall relative success rate boost for all user goal portions is on average 112\%, i.e. a move from 40\% to 85\% in absolute terms. For the test set, this difference is even larger, from 22 to 80\% absolute success rate, resulting in 263\% relative boost. 

These results show that transferring the knowledge from the source domain, we boost the target domain performance in data constrained regimes.

\subsection{Faster Learning}
\label{sec:warmstart}

In a second round of experiments, we study the effects of the transfer learning in the absence and in combination with the warm-starting phase. 
As warm starting requires additional labeled data, removing it further reduces the amount of labeled data needed. We also show that the two methods are compatible, leading to very good joint results.

\begin{figure}[h!]
\centering
\begin{subfigure}[t]{.5\textwidth}
\centering
\includegraphics[width=.49\textwidth]{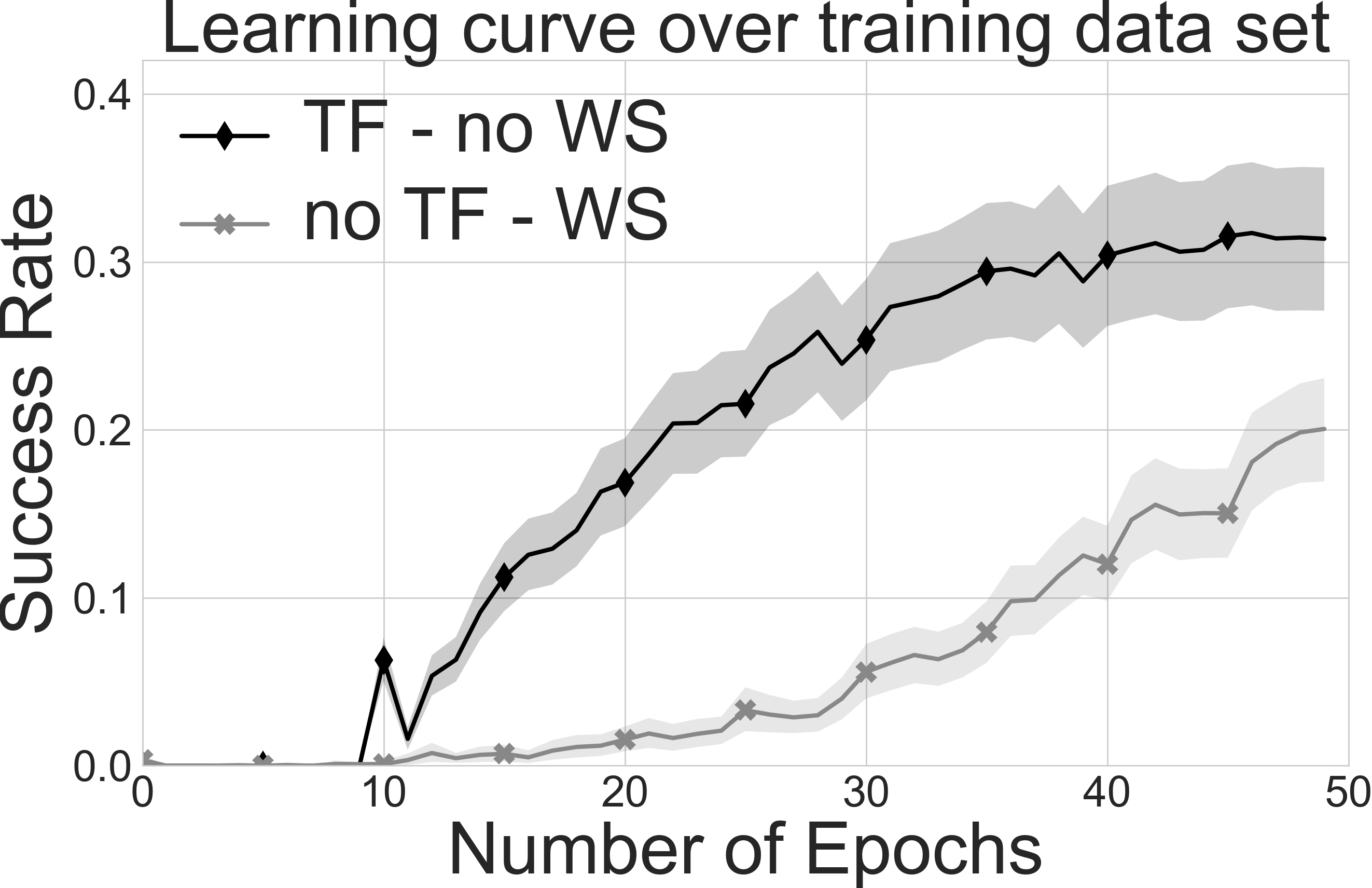}
\includegraphics[width=.49\textwidth]{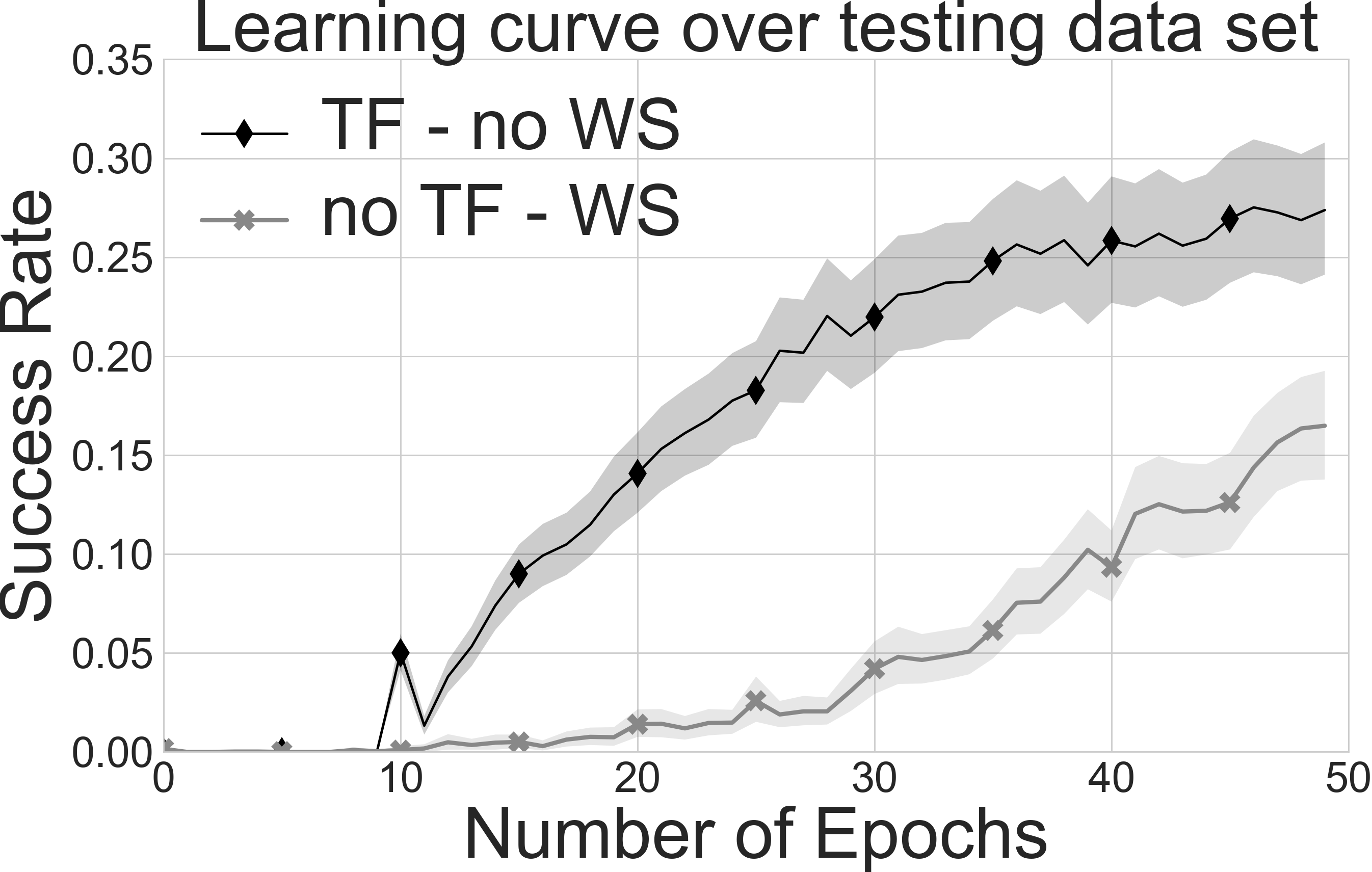}
\caption{Restaurant Booking with pre-training on Movie Booking domain}
\label{subfig:no_warm_up_rest_booking}
\end{subfigure}\\%
\begin{subfigure}[b]{.49\textwidth}
\centering
\includegraphics[width=.49\textwidth]{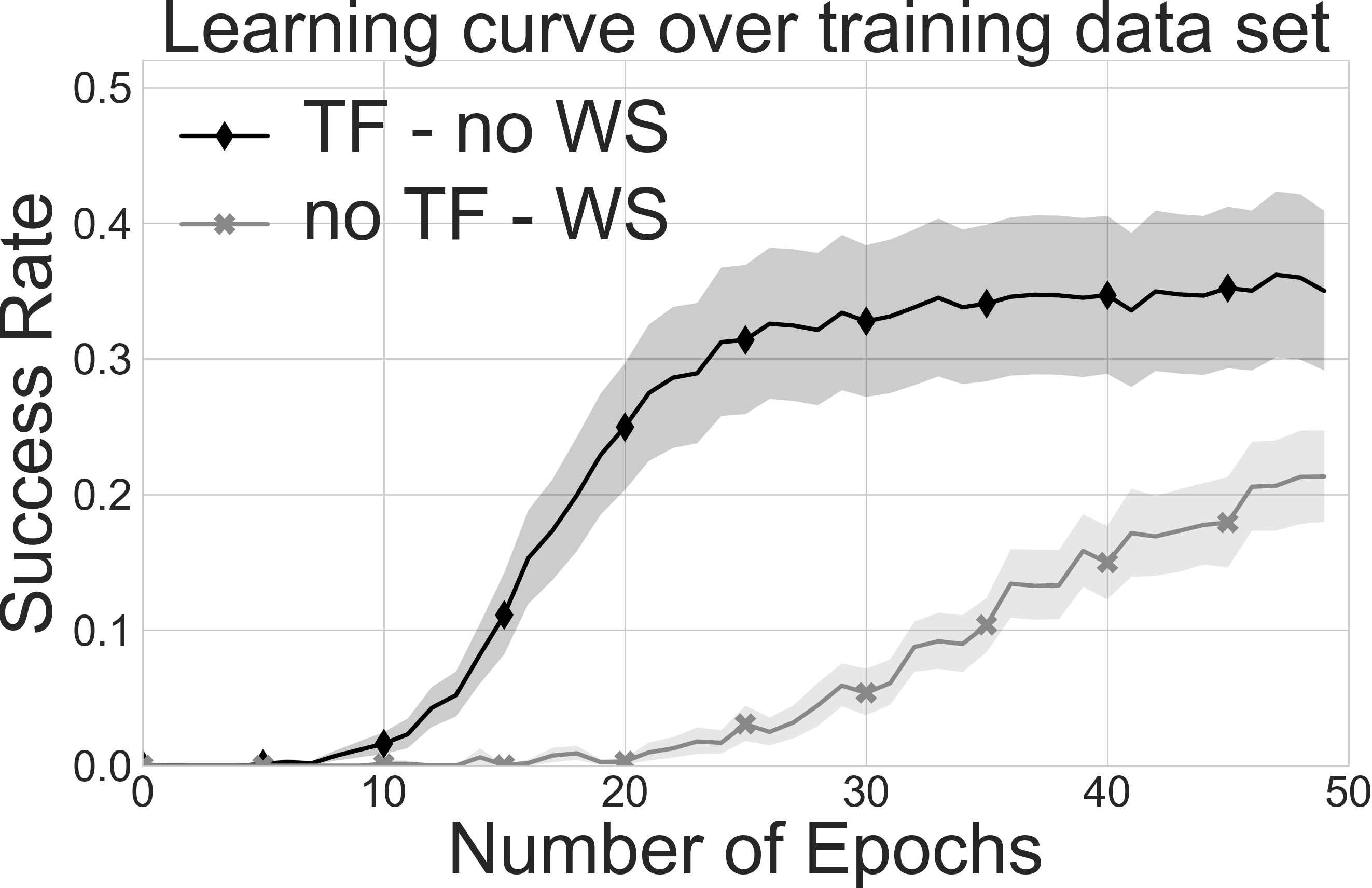}
\includegraphics[width=.5\textwidth]{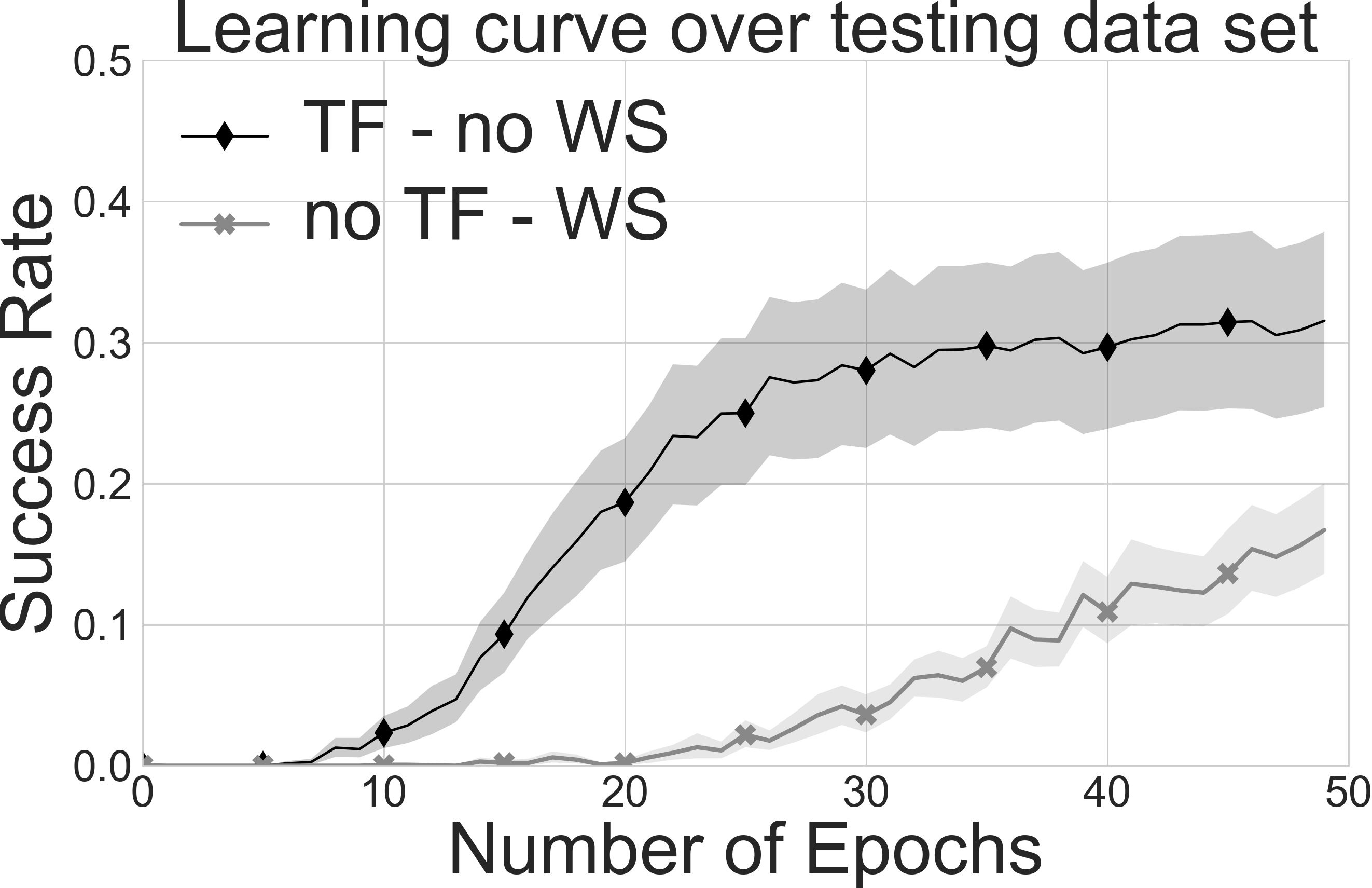}
\caption{Tourist Info with pre-training on Restaurant Booking domain}
\label{subfig:no_warm_up_tourist_info}
\end{subfigure}
\caption{Average training and testing success rates with 95\% confidence, for 100 runs over the number of epochs, for both models: with and without transfer learning (TF). The model with transfer learning is not warm-started (WS).}
\label{fig:no_warming_up_learning_curve}
\end{figure}

We report the training and testing learning curves (success rate over the number of training epochs), such that we use the full dataset of 120 training user goals and the test set of 32 user goals. We repeat the same process 100 times and report the results with a 95\% confidence interval.
The performances in the first case of domain overlapping are shown in Figure \ref{subfig:no_warm_up_rest_booking}, while for the other case of domain extension, in Figure \ref{subfig:no_warm_up_tourist_info}. The bot using transfer learning, but no warm-starting, shows better learning performances than the warm-started model without transfer learning. Transfer learning is thus a viable alternative to warm starting.

\begin{figure}[h!]
\centering
\begin{subfigure}[t]{.5\textwidth}
\centering
\includegraphics[width=.5\textwidth]{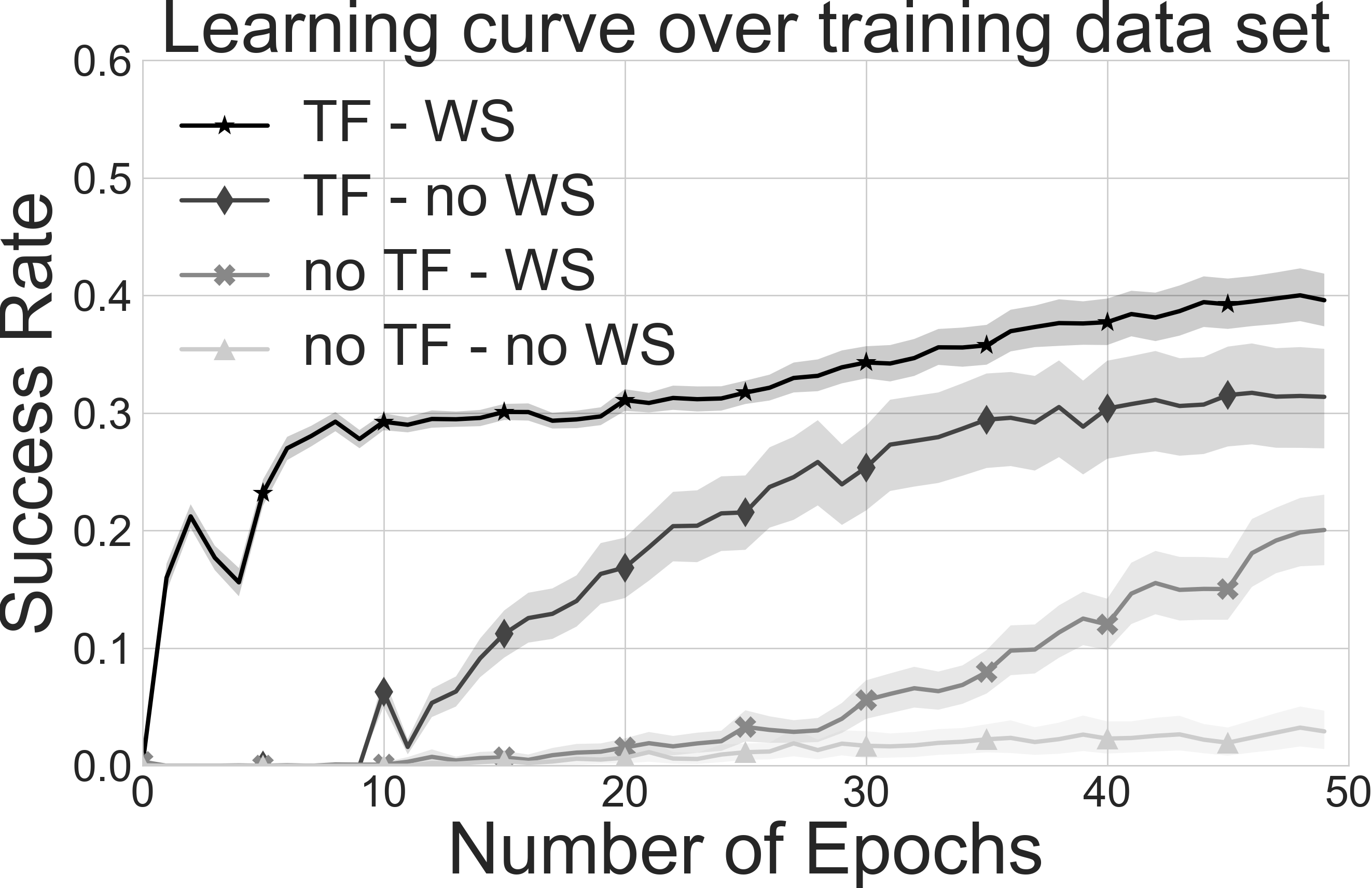}
\includegraphics[width=.49\textwidth]{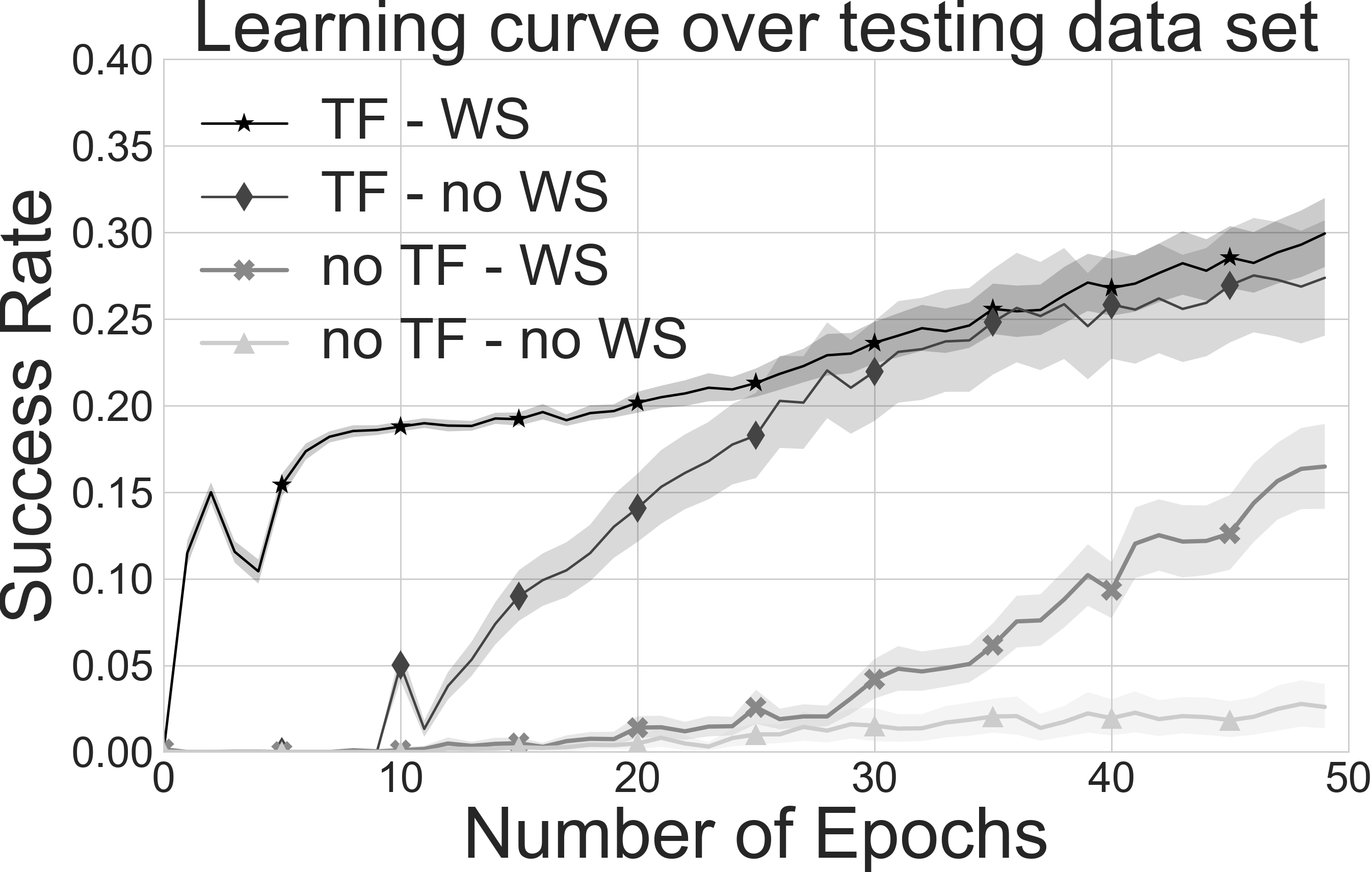}
\caption{Restaurant Booking with pre-training on Movie Booking domain}
\label{subfig:all_cases_rest_booking}
\end{subfigure}\\%
\begin{subfigure}[b]{.5\textwidth}
\centering
\includegraphics[width=.49\textwidth]{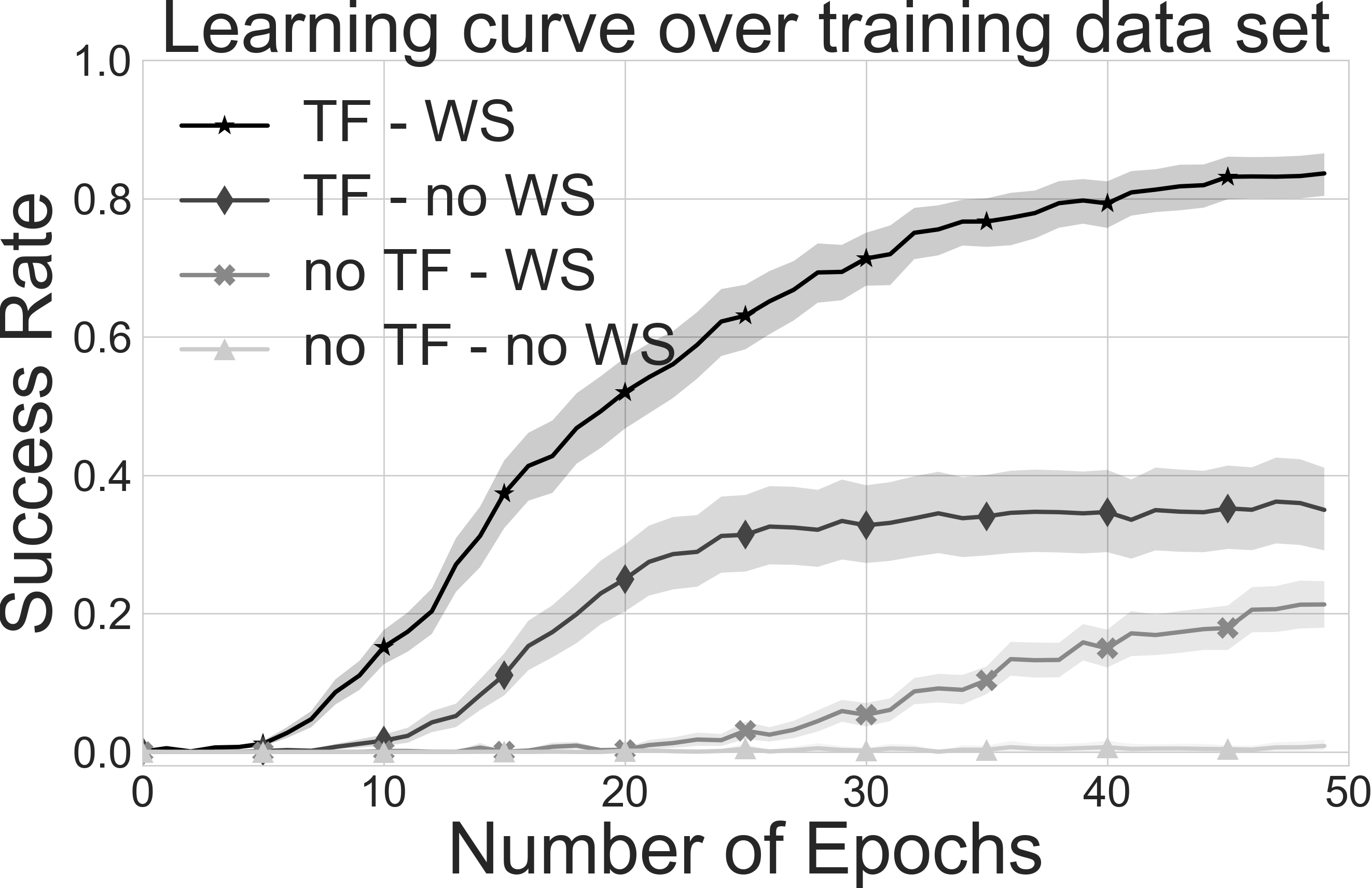}
\includegraphics[width=.5\textwidth]{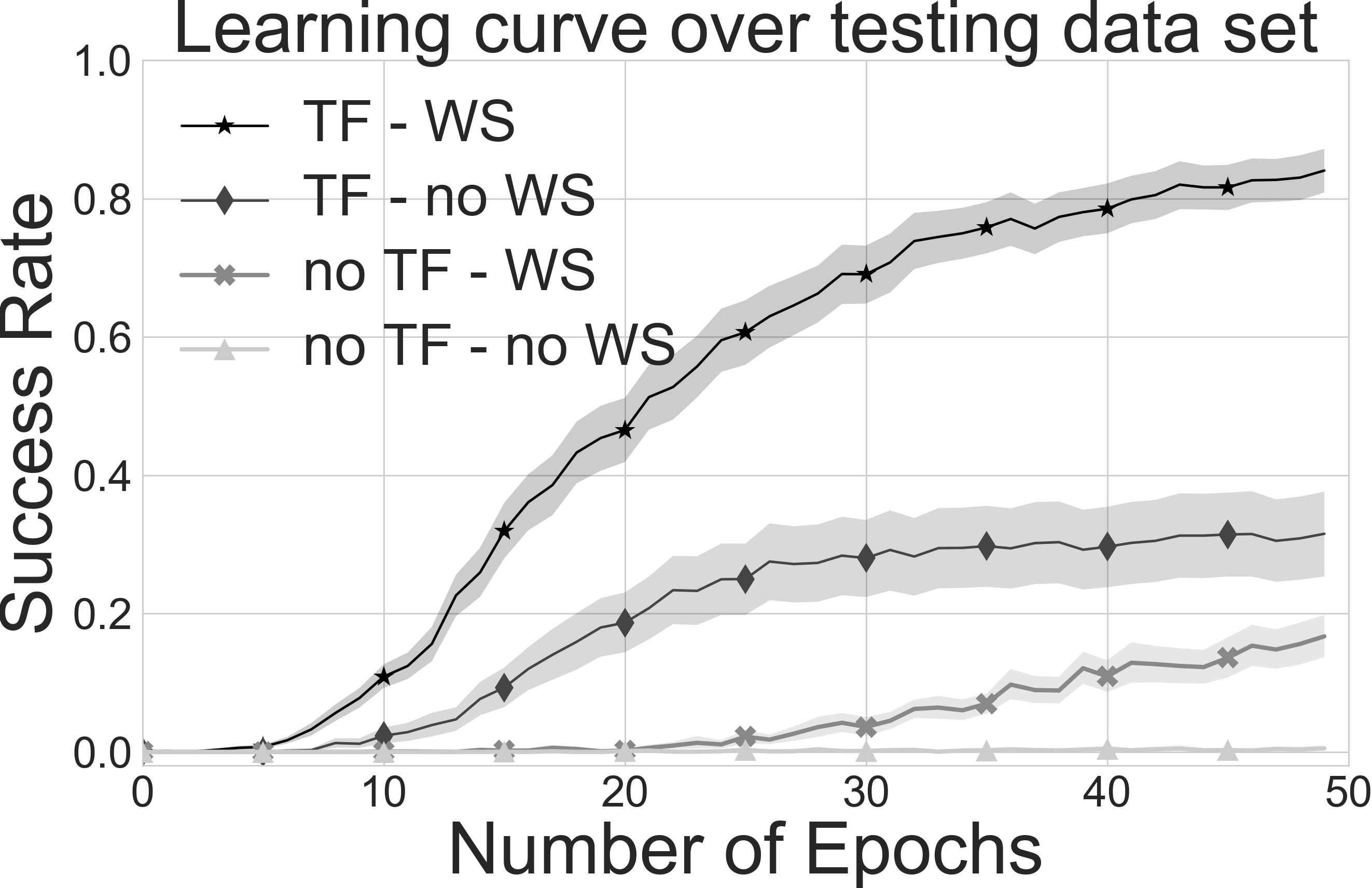}
\caption{Tourist Info with pre-training on Restaurant Booking domain}
\label{subfig:all_cases_tourist_info}
\end{subfigure}
\caption{Success rates for all model combinations - with and without Transfer Learning (TF), with and without Warm Starting (WS).}
\label{fig:all_cases_learning_curve}
\end{figure}

However, models based on transfer learning have a significant variance, as the learning is progressing. This happens because in many experiment runs the success rate over all epochs is 0. In those cases, the agent does not find an optimal way to learn the policy in the early stages of the training process. This results with filling its experience replay buffer mostly with negative experiences. Consequently, in the later stages, the agent is not able to recover. This makes a combination with warm starting desirable.

For convenience reasons, in Figure \ref{fig:all_cases_learning_curve} we show all possible cases of using and combining the transfer learning and warm-starting techniques. We can see that the model combines the two techniques performs the best by a wide margin. This leads to a conclusion that the transfer learning is complimentary to the warm-starting, such that their joint application brings the best outcomes.

\section{Conclusion}
\label{sec:conclusion}

In this paper, we show that the transfer learning technique can be successfully applied to boost the performances of the Reinforcement Learning-based Goal-Oriented Chatbots. We do this for two different use cases: $i)$ when the source and the target domain overlap, and $ii)$ when the target domain is an extension of the source domain. 

We show the advantages of the transfer learning in a low data regime for both cases. When a low number of user goals is available for training in the target domain, transfer learning makes up for the missing data. Even when the whole target domain training data is available, the transfer learning benefits are maintained, with the success rate increasing threefold.

We also demonstrate that the transfer knowledge can be a replacement of the warm-starting period in the agents or can be combined with it for best results. 

Last but not the least, we create and share two datasets for training Goal-Oriented Dialogue Systems in the domains of Restaurant Booking and Tourist Information.

\section*{Acknowledgements}
We would like to thank Patrick Thiran (EPFL), for sharing his valuable ideas and insights during the course of this research.

\bibliographystyle{named}
\bibliography{bibliography}

\end{document}